\documentclass{article}
\usepackage{ijcai22}

\usepackage{times}
\usepackage{soul}
\usepackage{url}
\usepackage[hidelinks]{hyperref}
\usepackage[utf8]{inputenc}
\usepackage[small]{caption}
\usepackage{graphicx}
\usepackage{amsmath}
\usepackage{amsthm}
\usepackage{booktabs}
\usepackage{algorithm}
\usepackage{algorithmic}
\usepackage{amssymb}

\title{Reconstruction Enhanced Multi-View Contrastive Learning for 
	
	Anomaly Detection on Attributed Networks  }

\author{
	Jiaqiang Zhang$^{1,2}$\and
	Senzhang Wang$^3$\And
	Songcan Chen$^{1,2}$\footnote{Corresponding author}\\
	\affiliations
	$^1$College of Computer Science and Technology, Nanjing University of Aeronautics and Astronautics\\
	$^2$MIIT Key Laboratory of Pattern Analysis and Machine Intelligence\\
	$^3$Central South University\\
	\emails
	\{zhangjq, s.chen\}@nuaa.edu.cn, 
	szwang@csu.edu.cn}

\begin{document}
	
	\maketitle
	
	\begin{abstract}Detecting abnormal nodes from attributed networks is of great importance in many real applications, such as financial fraud detection and cyber security. 
		This task is challenging due to both the complex interactions between the anomalous nodes with other counterparts and their inconsistency in terms of attributes.     
		This paper proposes a self-supervised learning framework that jointly optimizes a multi-view contrastive learning-based module and an attribute reconstruction-based module to more accurately detect anomalies on attributed networks.
		Specifically, two contrastive learning views are firstly established, which allow the model to better encode rich local and global information related to the abnormality.
		Motivated by the attribute consistency principle between neighboring nodes, a masked autoencoder-based reconstruction module is also introduced to identify the nodes which have large reconstruction errors, then are regarded as anomalies.  
		Finally, the two complementary modules are integrated for more accurately detecting the anomalous nodes.
		Extensive experiments conducted on five benchmark datasets show our model outperforms current state-of-the-art models.
	\end{abstract}
	
	\section{Introduction}
	Attributed networks are ubiquitous in many real-world scenarios, such as social networks and citation networks. Recently, anomaly detection on attributed networks is of great significance in many security-related applications, such as social spam detection \cite{jain2019spam}, financial fraud detection \cite{wang2019semi} and network intrusion detection \cite{shone2018deep}, and has attracted rising research interest.
	Though it has been extensively studied, detecting anomalies on attributed networks is still a challenging task \cite{ding2019deep}. One major challenge of this problem is that the abnormal patterns of nodes are related to not only their interactions with other nodes on the topological structure, but also their inconsistency in terms of node attributes \cite{ding2019deep,peng2018anomalous}.  
	
	The early anomaly detection techniques, such as matrix factorization \cite{li2017radar} and OC-SVM \cite{erfani2016high} have been widely used in many applications. A primary limitation of such methods is that they largely rely on feature engineering constructed by domain experts. 
	Recently, deep learning techniques, especially Graph Neural Networks, have been widely used in various graph mining tasks (e.g. link prediction and node classification), and have achieved considerable performance gains. GNN-based techniques \cite{ding2021inductive} have also been adopted to anomaly detection which aim to learn the anomaly-aware node representations. 
	Due to the high cost of obtaining anomaly samples, anomaly detection is usually performed in an unsupervised manner. 
	\cite{ding2019deep} proposed an unsupervised autoenconder-based method to detect the abnormal patterns effectively, which can capture the abnormal nodes according to the reconstruction errors. 
	\cite{liu2021anomaly} for the first time proposed a contrastive learning based framework for anomaly detection on graphs. Their model takes the local contextual information as supervision signals, and learns the representative features from the node-subgraph instance pairs. The final discriminative scores are used to detect abnormal nodes.
	\cite{jin2021anemone} performed patch- and context-level contrastive learning via two GNN-based models to further improve the performance.    
	
	Although considerable research efforts have been devoted to this problem recently, we argue that the current approaches still have shortcomings due to the following reasons.
	First, local and global structural information are not well utilized and integrated. Existing SOTA approach
	\cite{liu2021anomaly} typically constructed contrastive instance pair based on node-subgraph to efficiently focus on the local information of a node for anomaly detection. They  adopted one-layer GCN to extract the local structure information from the one-hop neighbors, while the global structure information beyond one-hop neighbors cannot be effectively captured and utilized. Previous studies show that higher-order neighboring information is also beneficial for graph mining tasks. \cite{ding2019deep} proposed to use the multi-layer GCN to capture the high-order interactions for anomaly detection on graphs, but the local and global structure information is modeled uniformly and coupled together, leading to information redundancy in feature representation learning.
	Meanwhile, GCN can be seen as a special form of low-pass filter, which has the effectiveness in smoothing signals \cite{bo2021beyond}. Stacking multiple layers of GCN will more likely smooth out the abnormal signals.
	Therefore, a more effective and decoupled local-global structure information extraction and integration method is needed to provide high quality structure features.
	Second, how to effectively fuse the node attributes with network topological structures to further boost the detection performance is not well explored. Although GCN can handle both topological structure and attributes, the learned representations are not well suitable for anomaly detection under the unsupervised learning scenario \cite{liu2021anomaly}. Anomalous nodes usually present inconsistency in terms of node attributes with their neighboring nodes, which provides us with additional self-supervised signals. How to fully utilize such signals and build an attribute reconstruction model based on the attribute consistency principle to further improve the anomaly detection performance is not well studied. 
	
	To alleviate the above drawbacks, we present a \underline{\textbf{Sub}}graph-based multi-view self-supervised framework which contains a \underline{\textbf{C}}ontrastive-based module and a \underline{\textbf{R}}econstruction-based module (\textbf{Sub-CR} for convenience).
	Sub-CR can separate the local information from the global information instead of directly fusing them together, and alleviate the risk of smoothing out abnormal signals in GNN. 
	Specifically, the original graph is first augmented by both graph diffusion and subgraph sampling to obtain local and global view subgraphs for each node. 
	Then a two-view (intra- and inter-view) contrastive learning module is proposed. The intra-view contrastive learning aims to maximize the agreement between the target node and its subgraph-level representations of both views, here the agreement can be quantified as a discriminative score.
	The inter-view contrastive learning aims to make the discriminative score of the two views closer, which allows the model to encode both local and global information.
	In order to further utilize the self-supervised signals of the attribute consistency between the neighboring nodes, 
	a masked autoencoder is also introduced to reconstruct the original attributes of the target node based on its neighboring nodes in both views. Finally, the two complementary modules are integrated for anomalous node detection.
	Our main contributions are summarized as follows:
	
	$ \bullet $ A novel self-supervised learning framework is proposed to address the problem of anomaly detection on attributed networks, which can effectively integrates contrastive learning- based and attribute reconstruction-based models.
	
	$ \bullet $ A multi-view learning model is proposed. Local and global information
	are modeled separately first and then integrated together, which can provide high quality features for detecting anomalous nodes.
	
	$ \bullet $ We conduct extensive experiments on five benchmark datasets for model evaluation. The results verify the effectiveness of our proposal and it outperforms existing state-of-the-art methods.
	\section{Related Work}
	\subsection{Anomaly Detection on Attributed Networks }
	The early work \cite{perozzi2016scalable} proposed a quality measure called normality, which utilized structure and attributes together on attributed networks. 
	\cite{li2017radar} proposed a learning framework to describe the residual of attributes and their consistency with network structure to detect anomalies in attributed networks. 
	With the rapid development of deep neural networks, researchers also tried to adopt deep learning techniques to detect anomalies in attributed networks. \cite{ding2019deep} proposed to use GCN and autoencoder to measure the reconstruction error of nodes from the perspective of structure and attributes to identify anomalies, which can alleviate the network sparsity and data nonlinearity issues. \cite{yu2018netwalk} adopted a clustering-based method through the learned dynamic network representation to dynamically identify network anomalies based on clustering in an incremental manner. \cite{li2019specae} proposed SpecAE to detect global anomalies and community anomalies via a density estimation approach.
	\cite{liu2021anomaly,jin2021anemone} proposed a contrastive self-supervised learning framework for graph anomaly detection, effectively capturing the relationship between nodes and their subgraphs to capture the anomalies.
	
	\subsection{Graph Contrastive Leaning}
	Recently, contrastive learning as an effective mean of self-supervised learning \cite{liu2021self}, has gained rising research interest, which aims to learn representations from supervised signals derived from the data itself. The core idea is that it considers two augmented views of the same data instance as positive sampler to be pulled closer, and all other instances are considered as the negative samplers to be pushed farther apart. Due to the great success of self-supervised learning on images \cite{chen2020simple}, contrastive learning methods are also adopted to various graph learning tasks. \cite{velickovic2019deep} aimed to learn the node representations by maximizing the mutual information  between the patch representation and the corresponding high-level graph summary. \cite{peng2020graph} directly maximized the mutual information between the input and output of the graph neural encoder according to the node features and topology to improve the DGI. \cite{zhu2020deep} proposed to generate two data views and then pull the representation of the same node in the two views closer, push the representation of all other nodes apart. \cite{zhu2021graph} proposed a contrastive framework for unsupervised graph representation learning with adaptive data augmentation.

	\section{Problem Formulation}
	In this paper, for the convenience of presentation, the calligraphic fonts (e.g. $\mathcal{G}$) represent sets, bold lowercase letters and bold uppercase letters represent vectors (e.g. \textbf{x}) and matrices (e.g. \textbf{X}), respectively. The terminology and problem definitions are given as follows. Let $\mathcal{G}=(\mathcal{V},\mathcal{E},\textbf{X})$ denote an attributed network , where $\mathcal{V}=$$ \left\{v_{1},...,v_{N}\right\} $ represents the set of nodes, $ N $ is the number of the nodes in the graph, $ \mathcal{E} $ is the set of edges, $ \textbf{X} $ $ \in $ $ \mathbb{R}^{N \times F} $  is the attribute matrix, where $ \textbf{x}_{i} \in \mathbb{R}^{F} (i=1,...,N) $ is the attributes for the $ i^{th} $ node, and $ \textbf{A} $$ \in \mathbb{R}^{N \times N} $ is the adjacency matrix.
	\begin{figure*}
		\begin{center}
			\includegraphics[scale=0.57]{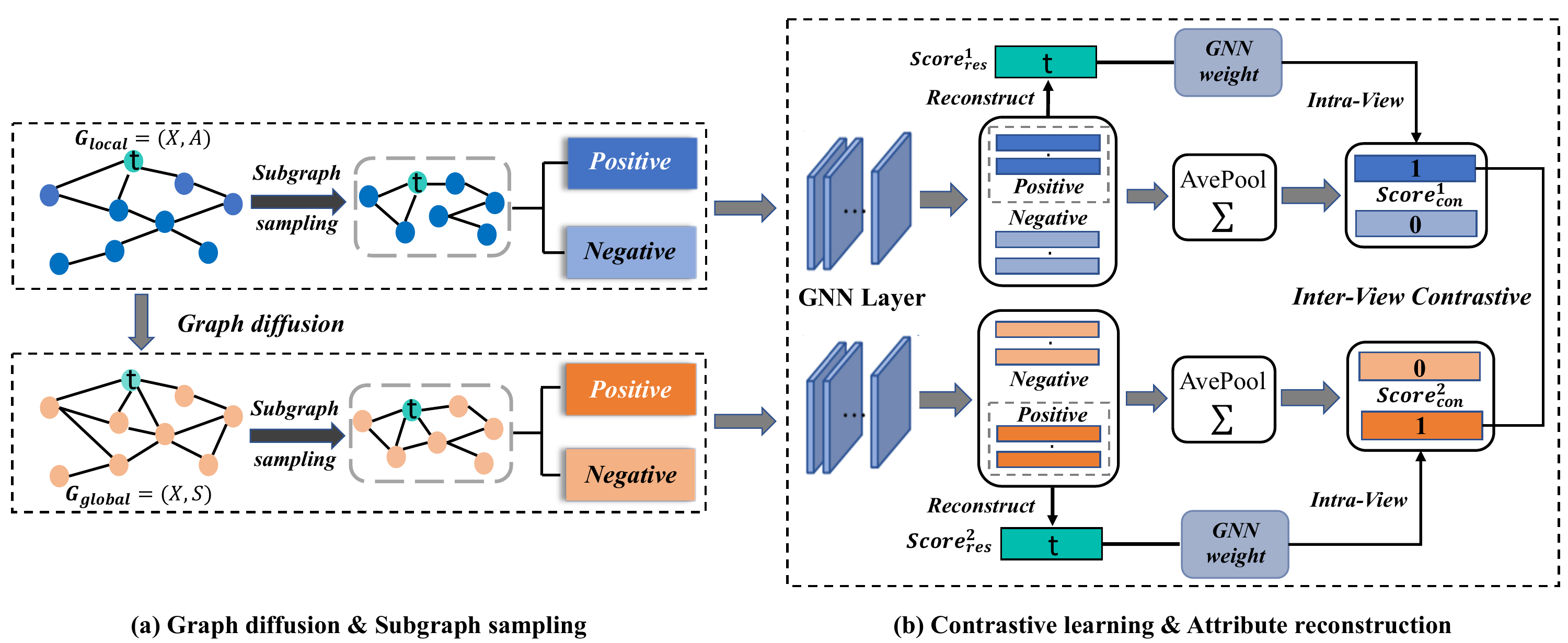}
			\caption{The framework of the proposed model.}
			\label{model}
		\end{center}	
	\end{figure*}		
	\paragraph{Problem Definition.} Given an attributed network $\mathcal{G}=(\mathcal{V},\mathcal{E},\textbf{X})$ whose adjacency matrix is $ \textbf{A} $, we aim to learn an anomaly score function $ score(\cdot) $ to measure the degree of the abnormality for each node in $\mathcal{G} $.

	\section{Methodology}
	
	As shown in Figure.\ref{model}, the proposed Sub-CR contains a contrastive learning-based module and an attribute reconstruction-based module. We first construct two types of data augmentations: the subgraph sampling (left upper part in Figure.\ref{model}) and the graph diffusion (left lower part in Figure.\ref{model}), to get local and global view subgraphs for each node. For each view, both the subgraph and the target node are input into a GNN encoder to learn the embedding as shown in the middle part in Figure.\ref{model}. Then we establish the intra- and inter-view contrastive learning (right part in Figure.\ref{model}) to encode local and global structure information. Meanwhile, we further introduce the masked autoencoder to identify the anomalous nodes through the attribute reconstruction errors in two views. Finally, the two complementary modules are integrated to give a final anomaly score.

	\subsection{Multi-View Graph Generation with Graph Diffusion and Subgraph Sampling}
	The effectiveness of the contrastive learning paradigm largely relies on the choice of the pretext task and the data augmentations \cite{liu2021self}. We adopt the node-subgraph matching pattern based pretext task which is proved useful for the anomaly detection on attributed networks \cite{liu2021anomaly}. Meanwhile, the graph diffusion augmentation \cite{hassani2020contrastive} is introduced to get the global view of the structure, which allows each node to match with its local- and global-view subgraphs respectively. Next, we introduce how to generate the two views of graph. 
	\paragraph{Graph Diffusion.} We propose to use the graph diffusion as the first augmentation to obtain the global view graph. Graph diffusion can effectively capture the global structure information \cite{hassani2020contrastive,klicpera2019diffusion}. 
	This process is formulated as follows:
	\begin{equation}
		\textbf{S}=\sum_{k=0}^\infty \theta_k\textbf{T}^{k} \in \mathbb{R}^{N \times N} 
	\end{equation}
	where $ \theta_k $ is the weighting coefficient to control the proportion of local and global structure information and $ \textbf{T} \in \mathbb{R} ^{N \times N}$ denotes the generalized transition matrix to transfer the adjacency matrix. Note that $ \theta_k \in[0,1] $ and
	$ \sum_{k=0}^\infty \theta_k=1 $.
	In this paper, Personalized PageRank (PPR) \cite{page1999pagerank} is adopted to power the graph diffusion. Given the adjacency matrix $ \textbf{A} \in \mathbb{R}^{N \times N}$, the identity matrix $ {\textbf{I} }$ and its degree matrix $ \textbf{D} $, the transition matrix and the weight can be formulated respectively as
	$ \textbf{T}={\textbf{D}}^{-1/2}\textbf{A}\textbf{D}^{-1/2} $ and $ \theta_k =\alpha(1-\alpha)^{k}$. Then the graph diffusion $ \textbf{S}$ can be reformulated as:
	\begin{equation}
		\textbf{S}=\alpha({\textbf{I}}-(1-\alpha){\textbf{D}}^{-1/2}\textbf{A}\textbf{D}^{-1/2})^{-1} 
	\end{equation}
	where $ \alpha $ denotes the teleport probability in a random walk. The graph diffusion $\textbf{S}$ is the global view of the initial graph.
	{\paragraph{Subgraph Sampling.} We adopt random walk with restart (RWR) \cite{tong2006fast} 
		to sample the local subgraph. In both views, a subgraph with the size $P$ is sampled for each node. In particular, we sample nodes and their edges from the original view to form the local-view subgraph, and then select the exact nodes and edges from the other view to form the global-view subgraph. In each view, for node $ v_{i} $, its subgraph is regarded as its positive pair, and a subgraph corresponding to the other node is regarded as the negative pair. 
		
		\subsection{\textbf{Local \& Global Graph Contrastive Learning}}
		As mentioned before, the representation of a normal node is similar to that of the neighboring nodes in its subgraph while this does not hold for an abnormal node.
		Therefore, we define the intra-view contrastive learning in the local and the global views, which maximizes the agreement between the node and the corresponding subgraph-level representations. As shown in the right part of Figure.\ref{model}, for node $v_{i}$, the discriminative score of its positive pair should be close to 1, while the negative one should be close to 0. 
		Meanwhile, the inter-view contrastive learning is defined between the two views. It aims to encourage the discriminative score that the same node between its subgraph in two views to be close.
		\paragraph{Intra-View Contrastive Learning.} This contrastiveness is defined in each view. Taking the local view in Figure.\ref{model} as an example, we obtain the corresponding positive and negative subgraphs $\mathcal{G}_{i} $, $ \tilde{\mathcal{G}_{i}} $  for the target node $ v_{i} $.
		Note that in order to make the obtained representation more discriminative, the attributes of the target node in the subgraph are masked. Then the subgraph are fed into a shared GCN encoder to learn a low-dimensional representation, which can be formulated as follows:   
		\begin{equation}
			\textbf{H}_{i}^{l}=\phi(\hat{\textbf{D}}^{-1/2}_{i}  \textbf{A}_{i}^{'} \hat{\textbf{D}} ^{-1/2}_{i} \textbf{H}_{i}^{(l-1)} \textbf{W}^{(l-1)})
		\end{equation}
		where  $ \hat{\textbf{D}_{i}} \in \mathbb{R}^{N\times N} $ is the degree matrix of                     
		$ \textbf{A}_{i}^{'}=\textbf{A}_{i}+\textbf{I}_{N} $, $\textbf{A}_{i}^{'}$ is the subgraph adjacency matrix with self-loop, $ \textbf{I}_{N} $ is the identity matrix and $ \textbf{W}^{0} \in \mathbb{R}^{F \times d} $ is a learnable weight matrix. Since the target node is masked in the subgraph, we employ the weight matrix of GCN to map the features into the same embedding space. It can be formulated as:
		\begin{equation}
			\textbf{h}_{i}^{l}=\phi(\textbf{h}_{i}^{l-1} \textbf{W}^{l-1})
		\end{equation}
		where $ {h}_{i}^{0}=\textbf{x}_{i}$, $ \phi $ is the activation function such as ReLU. So we obtain the representations $\textbf{H}_{i}  $  and  $ \tilde{\textbf{H}_{i} } $ for the two-view subgraphs $\mathcal{G}_{i} $, $ \tilde{\mathcal{G}_{i}} $, $\textbf{h}_{i} $ for the target node $ v_{i} $. In order to apply the node-subgraph matching pattern pretext task (the representation of node and its corresponding subgraph is more consistent), we take the average pooling function as the readout module to obtain the subgraph-level embedding vector $ \textbf{e}_{i} $: 
		\begin{equation}
			\textbf{e}_{i}=Readout(\textbf{H}_{i})=\sum_{k=1}^{n_{i}}\frac{(\textbf{H}_{i})_{k}}{n_{i}}
		\end{equation}
		where $ n_{i} $ is the number of nodes in the subgraph $ \mathcal{G}_{i}$. So the discriminative scores for the positive pair can be defined as:
		\begin{equation}
			{s}_{i}=\sigma (\textbf{h}_{i} \textbf{W}_{s} \textbf{e}_{i}^{T})
		\end{equation}
		where $ \textbf{W}_{s} $ is a learnable matrix.  Similarly, we can calculate $\tilde{{s}_{i}}$ for the nagetive pair. Hence, the local view contrastive loss for the instance pair $ (v_{i}, \mathcal{G}_{i} $, $ \tilde{\mathcal{G}_{i}}) $ can be formulated as:
		\begin{equation}
			{\mathcal{L}}^{1}_{intra}(v_{i})=-\frac{1}{2}(log({s}_{i})+log(1- \tilde{{s}_{i}}))
		\end{equation}
		Besides, we can similarly calculate $\mathcal{L}_{intra}^{2}$ for the global view. Finally, by combining the two losses, we have the intra-view contrastive learning objective function defined below:
		\begin{equation}
			{\mathcal{L}}_{intra}=\frac{1}{2N}\sum_{i=1}^{N}(\mathcal{L}_{intra}^{1}(v_{i})+\mathcal{L}_{intra}^{2}(v_{i}))
		\end{equation}
		\paragraph{Inter-View Contrastive Learning.} This contrastiveness is defined between the two views. Following the core idea in contrastive learning, we make the discriminative scores from node-subgraph pair of the two views closer.  
		So the inter-view contrastive loss between the local- and global-view can be formulated as:
		\begin{equation}
			{\mathcal{L}}_{inter}=(||\textbf{s}_{1}-\textbf{s}_{2}||_{F}^{2})
		\end{equation}
		where $ \textbf{s}_{1} $, $ \textbf{s}_{2} $ are the vectors that consist of positive pair discriminative scores in the two views. By combining the intra- and the inter-view contrastive learning, the overall loss function of the multi-view contrastive learning module is:
		\begin{equation}
			{\mathcal{L}}_{con}={\mathcal{L}}_{inter}+{\mathcal{L}}_{intra}
		\end{equation}
		
		\subsection{Attribute Reconstruction Based on Neighbors}
		The self-supervised learning method based on masked autoencoders (\textit{MAE}) has been proved to be effective in many fields. For example, the BERT \cite{devlin2018bert}-based pre-training model has achieved remarkable results in many NLP tasks, and similar works are also proposed in CV \cite{he2021masked}. The basic idea of these methods is to mask a part of a sentence or a picture and then input the unmasked part into the encoder and decoder for predicting the masked part. 
		
		Motivated by the masked autoencoder method, we propose to design an attribute reconstruction based on neighbors module to boost the performance. We adopt an asymmetric design. The encoder operates on the nodes in the subgraph that are not masked. The latent representations of these nodes are concatenated as the input of the lightweight decoder to reconstruct the masked node's raw attributes. Taking the local view as an example, for the subgraph $ \mathcal{G}_i  $ of node $ v_i $, we can get the representation of $ \mathcal{G}_i  $ through the GCN encoder. The reconstruction loss of the local view can be defined as:
		\begin{equation}
			{\mathcal{L}}^{1}_{res}(v_{i})=||g(\textbf{Z}_{i})-x_{i}||^{2}
		\end{equation}
		where $ g(\cdot) $ is the multilayer perceptron, $  \textbf{Z}_{i} $ is the concatenation of neighboring nodes' representations in the subgraph. Similarly, we can calculate $\mathcal{L}_{res}^{2}$ for the global view.
		The overall loss function of this module can be formulated as:
		\begin{equation}
			{\mathcal{L}}_{res}=\frac{1}{2N}\sum_{i=1}^{N}(\mathcal{L}_{res}^{1}(v_{i})+\mathcal{L}_{res}^{2}(v_{i}))
		\end{equation}
		\subsection{Anomaly Score Inference}
		To jointly train the contrastive learning module and the attribute reconstruction module, we optimize the following objective function:
		\begin{equation}
			{\mathcal{L}}=\mathcal{L}_{con}+\gamma \mathcal{L}_{res}
		\end{equation}
		where  $ \gamma $ is a controlling parameter which balances the importance of the two modules.
		
		By minimizing the above objective function, we can compute the anomaly score of each node. Inspired by  \cite{liu2021anomaly}, taking the node $ v_{i} $ as an example, the anomaly score of the contrastive-based module can be calculated by:
		\begin{equation}
			{{{\textit{score}}}_{con}}(v_{i})=\frac{1}{2}[score_{con}^{1}(v_{i})+score_{con}^{2}(v_{i})]
		\end{equation}
		where $ score_{con}^{1}(v_{i})= s_{i}^{1-}-s_{i}^{1+} $, $ score_{con}^{2}(v_{i})= s_{i}^{2-}-s_{i}^{2+} $, $ s_{i}^{1+}$($ s_{i}^{2+}$) and $ s_{i}^{1-} $ ($s_{i}^{2-} $) are the local (global) view discriminantive scores of positive and negative pairs by Eq.$ (6) $.  
		The anomaly score from the reconstruction-based module is:
		\begin{equation}
			{{{\textit{score}}}_{res}}(v_{i})=\frac{1}{2} \sum_{k=1}^{2}score^{k}_{res}(v_{i})
		\end{equation}
		
		\begin{equation}
			score^{k}_{res}(v_{i})=||g(\textbf{Z}_{i}^{k})-x_{i}||_{2}^{2}, k=1, 2
		\end{equation}
		where $ \textbf{Z}_{i}^{1} $ and $ \textbf{Z}_{i}^{2}$ are the concatenated neighboring node representations in local and global subgraphs for node $ v_{i} $, respectively. The two scores are first normalized, and then integrated by the following formula: 
		\begin{equation}
			{\textit{score}}(v_{i})=	{{{\textit{score}}}_{con}}(v_{i})+\gamma{{{\textit{score}}}_{res}}(v_{i})
		\end{equation}
		\section{Experiments}
		\subsection{Datasets}
		We evaluate the proposed model on five benchmark datasets that are widely used in anomaly detection on attributed networks \cite{liu2021anomaly,zheng2021generative}. These datasets include
		two social network datasets BlogCatalog and Flickr and three citation network datasets Cora, Citeseer, and Pubmed. Due to the shortage of ground truth anomalies, the anomalies are injected by the perturbation scheme \cite{ding2019deep}. The statistics of the datasets is shown in Table 1.
		
		\begin{table}[!t]
			\centering
			\small
			\resizebox{.9\columnwidth}{!}{
				\begin{tabular}{c|c|c|c|c}
					\toprule
					\bottomrule
					{Datasets} & \multicolumn{1}{c}{{Nodes}} & \multicolumn{1}{c}{{Edges}} & \multicolumn{1}{c}{{Features}} & {Anomalies} \\
					\midrule
					{BlogCatalog} & 5196  & 171743 & 8189  & 300 \\
					{Flickr} & 7575  & 239738 & 12407 & 450 \\
					{Cora} & 2708  & 5429  & 1433  & 150 \\
					{CiteSeer} & 3327  & 4732  & 3703  & 150 \\
					{Pubmed} & 19717 & 44338 & 500   & 600 \\
					\midrule
					\bottomrule
			\end{tabular}}%
			\caption{The statistics of the datasets.}
			\label{2}%
		\end{table}
		
		\begin{table*}[!t]
			\centering
			\small
			\setlength{\tabcolsep}{5.6mm}
			\begin{tabular}{c|cccccc}
				\toprule
				\bottomrule
				Methods  & Blogcatalog    & Flickr  & Cora  & Citeseer & Pubmed \\
				\midrule
				AMEN$ ^{[2016]} $  & 0.6392  & 0.6573 & 0.6266  & 0.6154  & 0.7713  \\
				Radar$ ^{[2017]} $  & 0.7401      & 0.7399  & 0.6587  & 0.6709  & 0.6233  \\
				ANOMALOUS $ ^{[2018]} $& 0.7237  & 0.7434  & 0.5770  & 0.6307  & 0.7316  \\
				DOMINANT$ ^{[2019]} $ & 0.7468  & 0.7442  & 0.8155  & 0.8251  & 0.8081  \\
				DGI $ ^{[2019]} $   & 0.5827  & 0.6237   & 0.7511  & 0.8293  & 0.6962  \\
				CoLA$ ^{ [2021]} $ & 0.7854  & 0.7513    & 0.8779  & 0.8968  & 0.9512  \\
				ANEMONE $^{ [2021] } $ & 0.8067  & 0.7637  & 0.9057  & 0.9189 & 0.9548 \\
				\midrule
				
				Sub-CR & \textbf{0.8141} & \textbf{0.7975}  & \textbf{0.9132} & \textbf{0.9303} & \textbf{0.9709} \\
				\midrule
				\bottomrule
			\end{tabular}%
			\caption{The AUC values comparison on five benchmark datasets.}
			\label{tab:addlabel}%
		\end{table*}%
		
		\subsection{Experimental Settings}
		\paragraph{Baselines and evaluation metrics.} In the experiments, we compare the proposed model with seven baselines. 
		\cite{perozzi2016scalable} evaluates the attribute correlation of node to detect anomalies. Specifically, it analyzes the abnormality of each node from the ego-network point of view.
		\cite{li2017radar} is a residual analysis-based method. It detects anomalies whose behaviors are singularly different from the majority
		\cite{peng2018anomalous} proposes a novel joint framework to conduct attribute selection and anomaly detection.
		\cite{ding2019deep} conducts the investigation on the problem of anomaly detection on attributed networks.
		\cite{velickovic2019deep} is an approach for learning node representations. A bilinear function is provied to score abnormality.
		\cite{liu2021anomaly} is the first contrastive learning based method for graph anomaly detection.
		\cite{jin2021anemone} is a recent method which performes patch- and context-level contrastive learning for anomaly detection on attributed networks. 
		We use ROC and AUC to measure the performance.
		\paragraph{Implementation Details.} In Sub-CR, we set the node size $ P $ of subgraph to 4 for all datasets by considering both the efficiency and performance. Following \cite{liu2021anomaly}, we employ a one-layer GCN as the encoder, and the embedding dimension is set to 64. The model is optimized with the Adam optimizer during training. The bacth size is set to 300 for all datasets. The learning rate is set to 0.001 for Cora, Citeseer, Pubmed and Flickr, set to 0.003 for BlogCatalog.  We train the model 400 epochs for BlogCatalog and Flickr, and 100 epochs for Cora, Citeseer and Pubmed. The parameter $\gamma$ is set to $ 0.6 $ for BlogCatalog, Flickr, Cora and Citeseer and $ 0.4 $ for Pubmed. In the inference phase, we average the anomaly scores in 
		Eq. (17) over 300 rounds to get the final anomaly score for each node. 
		\begin{figure}
			\centering
			\includegraphics[scale=0.28]{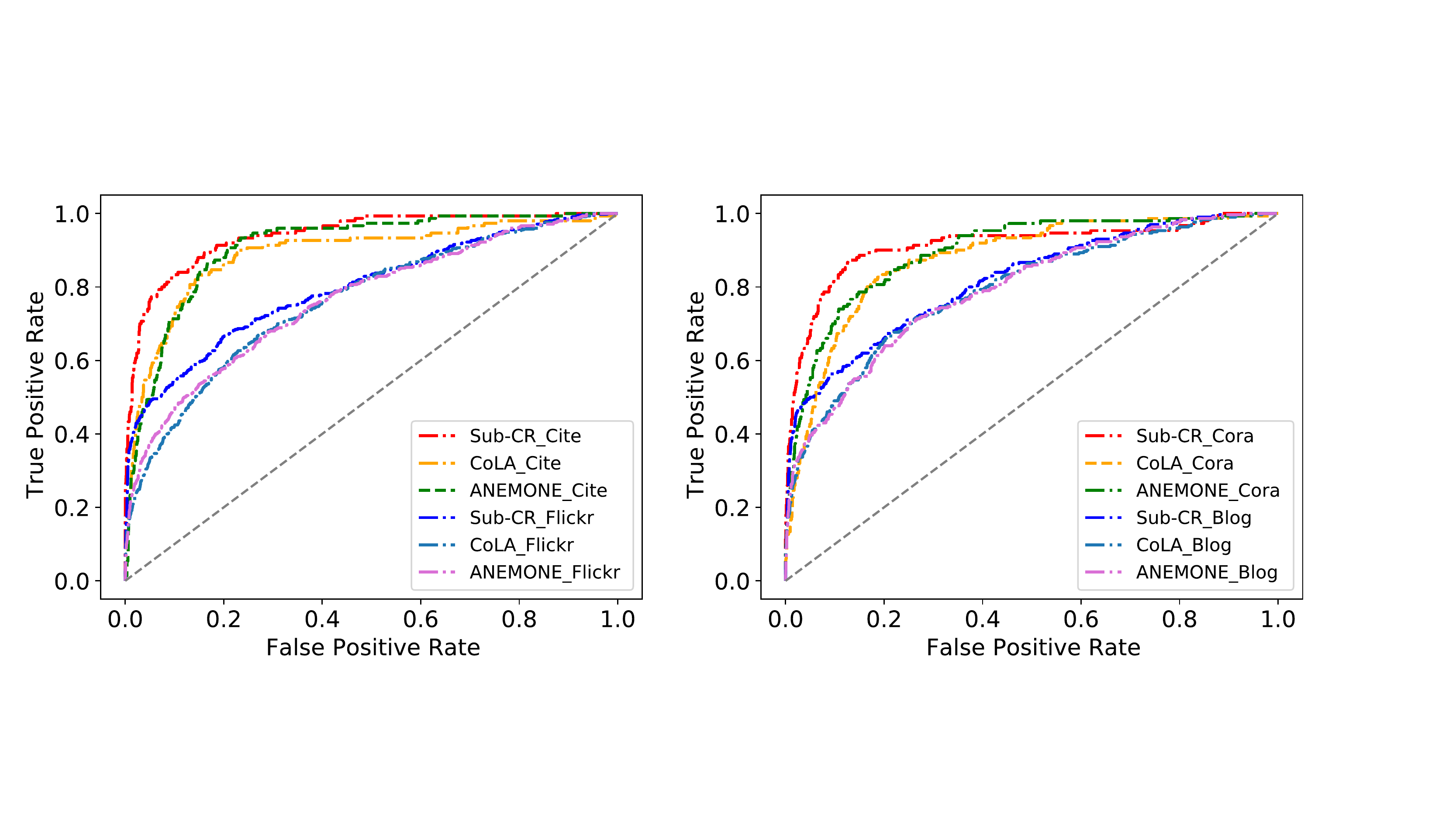}
			\caption{ROC curves on four benchmark datasets.}
			\label{roc}
		\end{figure}
		
		\subsection{Result and Analysis}
		Table 2 shows the comparison result of these methods in terms of AUC value. Due to space limitation, Figure.\ref{roc} shows the ROC curves on Cora, BlogCatalog, Citeseer, and Flickr with two competitive baselines.
		As shown in the table, one can first see that our proposed model is superior to its counterparts on five benchmark datasets, which demonstrates the effectiveness of Sub-CR. 
		To be specific, Sub-CR improves the performance by 3.38\% and 1.61\% over the best baseline ANEMONE on Flickr and Pubmed datasets, respectively. 
		In details, the shallow methods including AMEN, Radar and ANOMALOUS perform worse than other models due to their less effectiveness in modeling the high-dimension node attributes and complex tological structures. The DOMINANT and DGI also do not show the competitive performance because they focus on modeling the whole graph structure instead of directly explore the abnormal patterns. DGI takes the node vs full-graph intstance pair contrastive, and DOMINANT aims to reconstruct the whole graph structure or attributes for each node, which do not decouple local and global information for anomaly detection. 
		CoLA and ANEMONE achieve the suboptimal performance due to the consideration of sorrounding substructures. But they cannot fully catch the high-order structure information and ignore the self-supervised signal of the original attribute reconstruction error. The superior performence of Sub-CR verifies the effectiveness of intergating the multi-view contrastive-based and reconstruction-based modules, which can decouple local and global information to better capture the attribution and the topological structure of nodes for anomaly detection.
		\begin{table}[!t]
			\centering
			\small
			\resizebox{.99\columnwidth}{!}{
				\begin{tabular}{c|ccccc}
					\toprule
					\bottomrule
					& BlogCatalog & Flickr & Cora  & CiteSeer & Pubmed \\
					\midrule
					Sub-R & 0.7943  & 0.7609  & 0.9002  & 0.9017  & 0.9553  \\
					Sub-C & 0.7460  & 0.7434  & 0.8220  & 0.7892  & 0.8006  \\
					Sub-weight & 0.8083  & 0.7928  & 0.9041  & 0.9275  & 0.9491  \\
					Sub-global & 0.8090  & 0.7923  & 0.8975  & 0.9195  & 0.9625  \\
					\midrule
					\bottomrule
			\end{tabular}}
			\caption{The AUC values of ablation study.}
			\label{tab:}%
			\vspace{0.6em}
		\end{table}%
		
		\begin{figure}
			\centering
			\includegraphics[scale=0.42]{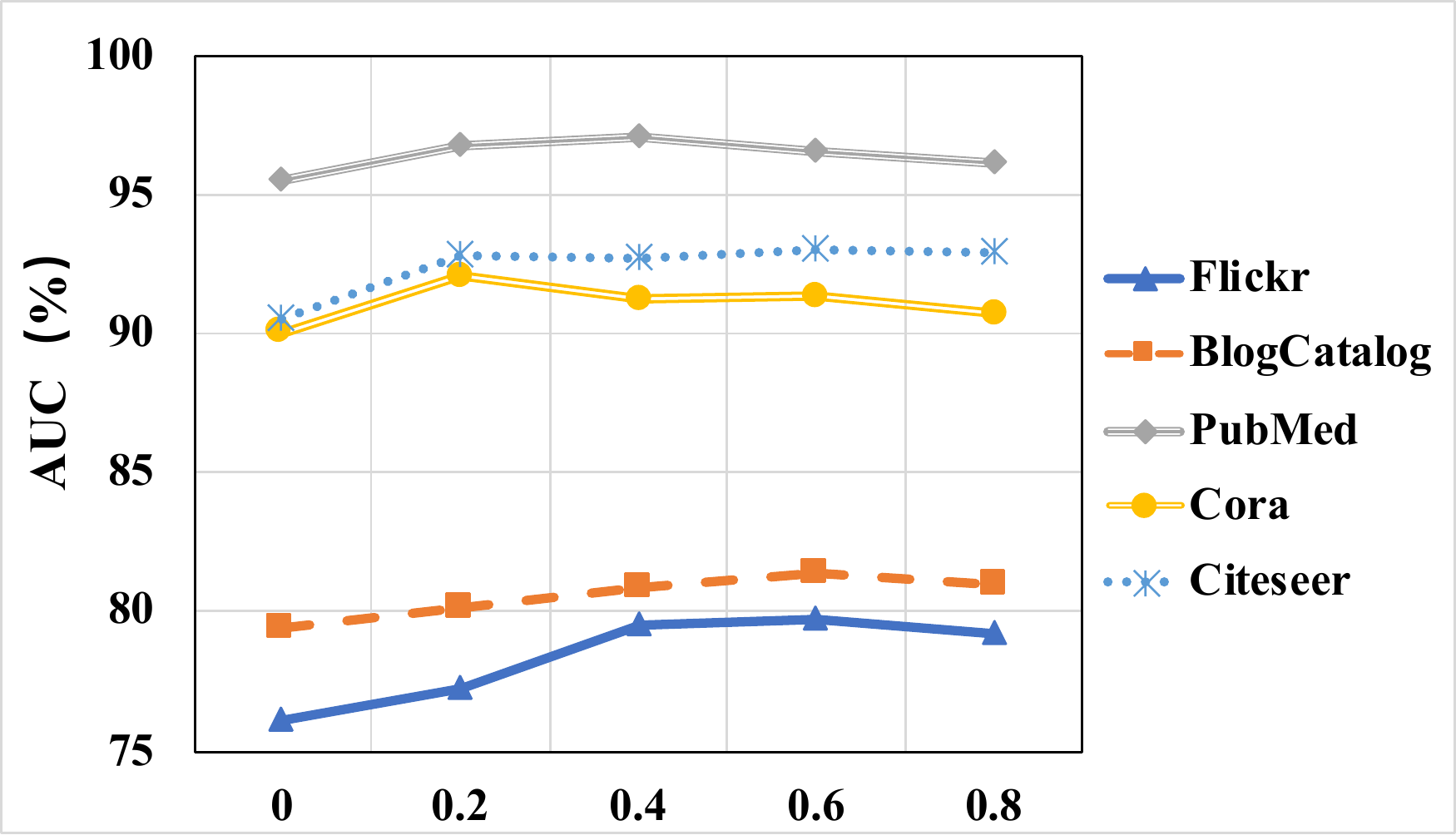}
			\caption{Performance with different $\gamma$. }
			\label{gama}
			
		\end{figure}
		
		\subsection{Ablation Study}
		We next conduct an ablation study to verify the effectiveness of each component in Sub-CR. Sub-R and Sub-C denote the model without reconstruction-based and contrastive-based module, respectively. The variants Sub-weight and Sub-global are defined as that we remove the balance weight ( $ \gamma $ is set to 1) and the global view, respectively.
		The results are presented in Table 3. One can see that Sub-CR outperforms all the variants consistently on all the datasets, which demonstrates the components are all helpful to the studied task. 
		Especially, Sub-CR outperforms Sub-C by 6.81\%, 5.41\%, 9.12\%, 14,11\%, 17.03\% on five datasets, respectively, which demonstrates the contrastive-based module is crucially important to the model.
		The self-supervised signal of the matching pattern on the node-subgraph pair is more effective for capturing anomalies. 
		As for the remaining three components, Sub-R performs worse than Sub-weight and Sub-global on BlogCatalog, Flickr, and CiteSeer datasets. On the Cora, the component of the global view is more important. 
		Sub-weight has a more significant effect on Pubmed.
		The reason may be that the effect of different modules on different datasets varies due to the different data characteristics.
		
		\begin{figure}
			\centering
			\includegraphics[scale=0.32]{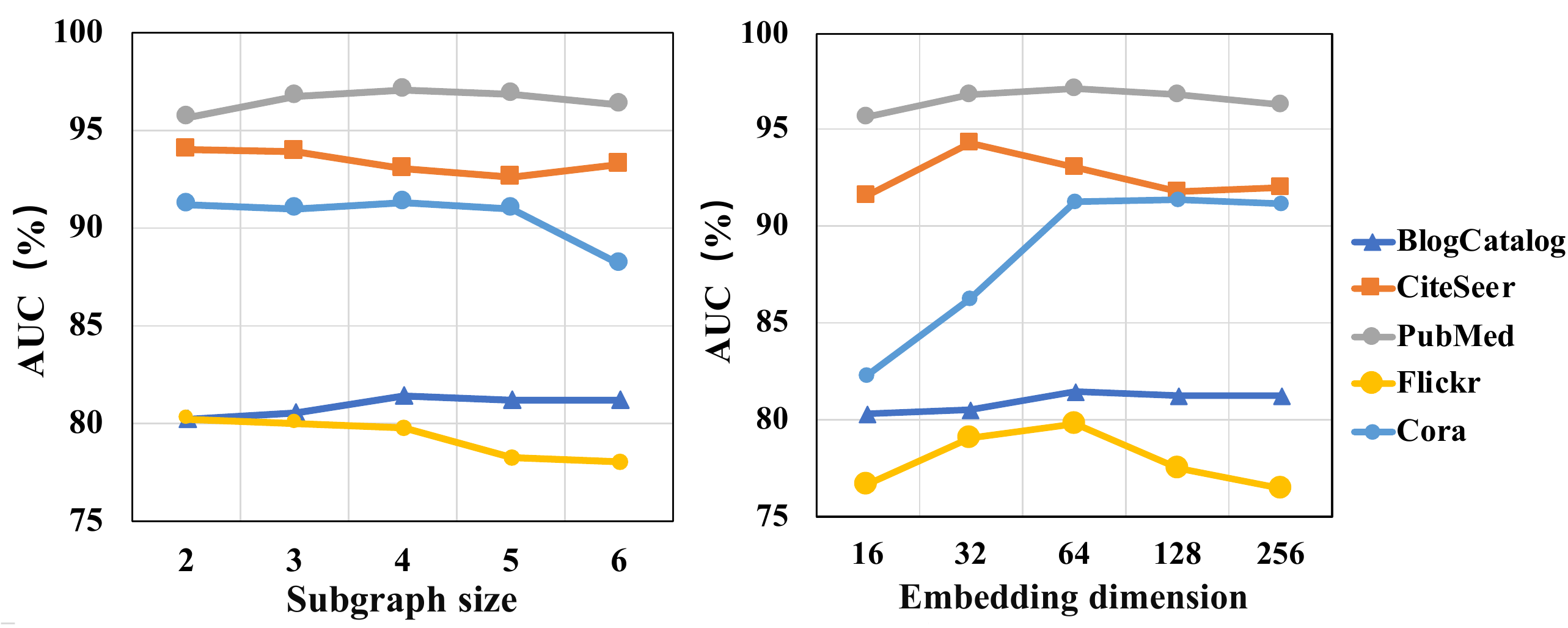}
			\caption{ Performance with different subgraph sizes and embedding dimensions.}
			\label{para}
		\end{figure}

		\subsection{Parameter Study}
		We finally conduct the model sensitivity analysis on critical hyper-parameters in Sub-CR, which are the size $ P $ of subgraph, the embedding dimension in GCN and the balance factor $ \gamma $. Figure.\ref{gama} shows the effect of different $ \gamma $ values on the model performance. One can see that the most datasets can achieve the best result and not sensitive to the balance factor when $ \gamma \geq 0.4 $.
		Figure.\ref{para} shows the AUC values under different subgraph sizes and embedding dimensions. One can observe that the best performance for Flickr and Citeseer is achieved at $ P=2 $, and for BlogCatalog, Cora and PubMeb it is achieved at $ P=4 $.  When the size is too large, the subgraph will contain redundant information, which will hurt the performance. A suitable embedding dimension is $ d=64$ for most datasets as shown in the figure. A too small or large embedding dimension will both degrade the model performance.
		
		\section{Conclusion}
		In this paper, we propose a novel multi-view self-supervised learning framework for anomaly detection on attributed networks. The proposed contrastive learning-based module consists of two carefully designed contrastive views to better capture local and global structure information related to anomaly patterns. The attribute reconstruction module adopts the representation of those neighbors based on the subgraph to rebuild the raw attributes of the target node of the two views. Finally, the two complementary modules are integrated for more effective anomaly detection. Extensive experiments conducted over five benchmark datasets demonstrate the effectiveness of our proposed model.
		\section*{Acknowledgements}
		This work was supported by NSFC under granted No. 62076124, partially financially supported by the National Science and Technology Major Project (J2019-IV-0018-0086), NSFC (No. 62172443), and the Fundamental Research Funds for the Central Universities (No. NZ2020014).

		\bibliographystyle{named}
		\bibliography{ijcai22}

	\end{document}